\documentclass[doublecol]{epl2} 

\usepackage{graphicx} 
\usepackage{hyperref}
\usepackage{xcolor}
\usepackage{amsmath} 
\usepackage{amssymb} 
\usepackage{amsthm}

\title{On the class of coding optimality of human languages and the origins of Zipf's law}
\shorttitle{On the class of coding optimality of human languages}

\author{Ramon Ferrer-i-Cancho\footnote{E-mail: rferrericancho@cs.upc.edu (corresponding author).} \inst{1}}
\shortauthor{R. Ferrer-i-Cancho}

\institute{                    
  \inst{1} Quantitative, Mathematical and Computational Linguistics Research Group, Department of Computer Science, Universitat Politècnica de Catalunya, 08034 Barcelona, Catalonia (Spain)
}

\abstract{
Here we present a new class of optimality for coding systems. Members of that class are displaced linearly from optimal coding and thus exhibit Zipf's law, namely a power-law distribution of frequency ranks. Within that class, Zipf's law, the size-rank law and the size-probability law form a group-like structure. 
We identify human languages that are members of the class. All languages showing sufficient agreement with Zipf's law are potential members of the class. 
In contrast, there are communication systems in other species that cannot be members of that class for exhibiting an exponential distribution instead but dolphins and humpback whales might.
We provide a new insight into plots of frequency versus rank in double logarithmic scale. For any system, a straight line in that scale indicates that the lengths of optimal codes under non-singular coding and under uniquely decodable encoding are displaced by a linear function whose slope is the exponent of Zipf's law. For systems under compression and constrained to be uniquely decodable, such a straight line may indicate that the system is coding close to optimality. We provide support for the hypothesis that Zipf's law originates from compression and define testable conditions for the emergence of Zipf's law in compressing systems. 
}

\begin{document}

\maketitle

\newtheorem{property}{Property}
\newtheorem{definition}{Definition}

\section{Introduction}

Nearly 7000 languages are spoken on Earth \cite{Leben2018a}. As languages from the same linguistic family resemble each other for sharing a close common ancestor, a better proxy for the diversity of languages is the number of families. That number, estimated at about 430, is still a large number \cite{Leben2018a}. In spite of such diversity, languages exhibit statistical patterns called linguistic laws. One of laws with strongest support is Zipf's law of abbreviation, the tendency of more frequent words to be shorter \cite{Zipf1949a}, that is supported in 986 languages, namely 14\% of the languages spoken on earth \cite{Bentz2016a}. The strength of the law relies on the fact that it is found independently of the linguistic family, writing system (script), the unit (words or individual characters), modality (oral versus written), the unit of measurement (characters, syllables, strokes or duration) \cite{Sanada2008a,Bentz2016a,Wang2015a,Meylan2021a,Petrini2022b,Koshevoy2023a}. 
Furthermore, this law is being found in a growing number of species with some interesting exceptions \cite{Semple2021a}. 

The pervasiveness of the Zipf's law of abbreviation calls for an explanation.
Let us consider a system that produces a certain unit $i$, whose probability is $p(i)$ and whose magnitude is $l(i)$. Without loss of generality, we assume that the $p(i)'s$ are sorted non-increasingly, that is 
\begin{equation*}
p(1) \geq p(2) \geq ... \geq p(i) \geq p(i+1) \geq ...
\end{equation*}
The magnitude can be a discrete quantity (e.g., the number of characters, syllables or strokes making a word), duration) or a real number (e.g., the duration of a vocalization). 
In this setting, the average magnitude is
\begin{equation}
\left<l \right> = \sum_i p(i) l(i)
\label{eq:average_code_length}
\end{equation}
and an operational definition of the law of abbreviation is that the Kendall $\tau$ correlation between $p(i)$ and $l(i)$ is such that \cite{Semple2021a,Ferrer2019c}
\begin{equation}
\tau(p(i), l(i)) < 0.
\label{eq:law_of_abbreviation}
\end{equation} 
It has been demonstrated that, if $\left<l \right>$ is minimum, then the Kendall $\tau$ correlation between $p(i)$ and $l(i)$ cannot be positive, i.e. \cite{Ferrer2019c} $\tau(p(i), l(i)) \leq 0$,
which sheds light on the origins of the law of abbreviation.

\begin{table}
\caption{\label{tab:codes} Distinct encodings of numbers from 1 to 6 with an alphabet formed by ${a, b}$. }
\begin{tabular}{llll}
       & \multicolumn{3}{c}{Code} \\ 
       \cline{2-4}
  Unit & Non-singular & Not non-singular & Elias gamma \\ \hline
  $1$ & $aa$ & $aa$ & $b$ \\
  $2$ & $ab$ & $aa$ & $aba$ \\
  $3$ & $a$  & $a$  & $abb$ \\
  $4$ & $b$  & $b$  & $aabaa$ \\
  $5$ & $ba$ & $ba$ & $aabab$ \\
  $6$ & $bb$ & $bb$ & $aabba$
\end{tabular}
\end{table}

In standard information theory, one assumes that $l(i)$ is the length of the code assigned to unit $i$. The code is a string of characters from some alphabet of size $N$. The problem of compression consists of finding the codes whose length minimizes $\left<l \right>$. We use $\left<l \right>_{min}$ to refer to the minimum value of $\left<l \right>$.
If no constraint is imposed on the possible codes, the minima of $\left<l \right>$ are singular. If $l(i) = 0$ is allowed, $\left<l \right>$ is minimized by assigning the empty string to each unit. Then $\left<l \right>_{min}=0$ and information transmission is impossible with empty strings. If $l(i) > 0$ is required, $\left<l \right>$ is minimized by assigning a string of just one character to each unit. Then $\left<l \right>_{min}=1$ and $N$ has to be as large as the number of units for successful decoding, which, if possible, is eventually an overkill. For this reason, information theory is concerned about constraints on the mapping of units into codes, which are known as coding schemes \cite{Cover2006a}. 
In the non-singular coding scheme, every unit is assigned a distinct string. Suppose that $N = 2$ and the alphabet is $\{a, b\}$. Table \ref{tab:codes} shows a coding table that is non-singular and a coding table that is not non-singular because the string $aa$ is assigned to both unit 1 and unit 2. In the uniquely decodable coding scheme, a particular case of non-singular coding, the segmentation of a string resulting from the concatenation of codes must be unique given the coding table. For instance, $baba$ allows for many interpretations according to the non-singular encoding in Table \ref{tab:codes}: the segmentation $ba$, $ba$ implies that the units coded are $5$, $5$; the segmentation $b$, $a$, $b$, $a$ implies that the units coded are $4$, $3$, $4$, $3$, and so on. Therefore, the non-singular codes in Table \ref{tab:codes} do not satisfy the uniquely decodable coding scheme. In contrast, the Elias gamma encoding \cite{Elias1975a} in Table \ref{tab:codes} does so: the sequence $baba$ can only be segmented as $b$, $aba$ and then the units coded are certainly $1$, $2$.

Hereafter we assume $N > 1$. Under optimal non-singular coding, $\left<l \right>$ is minimized by \cite{Mandelbrot1966, Ferrer2019c}
\begin{equation}
l(i) = \left\lceil \log_N \left(\frac{N-1}{N}i + 1 \right) \right\rceil
\label{eq:hard_optimal_length_non_singular_coding}  
\end{equation}
while, under uniquely decodable encoding, $\left<l \right>$ is minimized by \cite{Shannon1948,Cover2006a}
\begin{equation}
l(i) = \left \lceil -log_N p_i \right \rceil.
\label{eq:hard_optimal_length_uniquely_decodable_encoding}  
\end{equation}
The previous equations are hard definitions that assume that the magnitude of a code must be discrete. 
For this reasons, we replace them by simpler soft versions, i.e.  
\begin{equation}
l(i) = \log_N i
\label{eq:soft_optimal_length_non_singular_coding}  
\end{equation}
for optimal non-singular coding and
\begin{equation}
l(i) = - \log_N p_i
\label{eq:soft_optimal_length_uniquely_decodable_encoding}  
\end{equation}
for optimal uniquely decodable encoding. These soft versions have already been used in previous research on optimal coding with continuous magnitudes \cite{Hernandez2019b,Torre2019a} and can be justified as follows. 
First, Eq. \ref{eq:hard_optimal_length_non_singular_coding} is well approximated by Eq. \ref{eq:soft_optimal_length_non_singular_coding} as $\frac{N-1}{N}$ converges quickly to 1 as $N$ increases. 
Second, the solution of the constrained minimization of $\left< l \right>$ by means of Lagrange multipliers produces directly Eq. \ref{eq:soft_optimal_length_non_singular_coding} and Eq.  \ref{eq:hard_optimal_length_uniquely_decodable_encoding} results {\em a posteriori} by imposing discreteness on $l(i)$ \cite{Cover2006a}. 
 
Here we present a class of coding systems that are close to optimal coding under the non-singular coding scheme and under the uniquely decodable scheme. We will show that languages within that class must exhibit Zipf's rank-frequency law, hereafter Zipf's law, that states that $p(i)$, the probability of the $i$-th most likely word in a text, can be approximated by \cite{Zipf1949a,Mehri2017a}
\begin{equation}
p(i) = c i^{-\alpha},
\label{eq:Zipfs_rank_frequency_law} 
\end{equation}
where $\alpha$ is the so-called exponent of the law and $c$ is a normalization factor that depends on $\alpha$.
We will show strong evidence of languages that belong to that class while we will show communication systems of other species that do not belong to that class.  
 
\section{A new class of coding optimality}

The coding efficiency, namely \cite{Borda2011a} 
\begin{equation*}
\eta = \frac{\left<l \right>_{min}}{\left<l \right>},
\end{equation*}
measures the closeness to optimal non-singular coding. 
We use $\left< l \right>_{min}^{ns}$ and $\left< l \right>_{min}^{ud}$ to refer to the minimum value of $\left< l \right>$ under non-singular ($ns$) coding and uniquely decodable ($ud$) encoding. Accordingly, we define 
$\eta^{ns} = \left<l \right>_{min}^{ns}/\left<l \right>$ and $\eta^{ud} = \left<l \right>_{min}^{ud}/\left<l \right>$. Eq. \ref{eq:average_code_length} with $l(i)$ as in Eq. \ref{eq:soft_optimal_length_non_singular_coding} yields the expected value of $\log_N i$, that is 
\begin{equation}
\left< l \right>_{min}^{ns} = \left<\log_N i \right> = \sum_i p(i) \log_N i.
\label{eq:average_log_rank}
\end{equation}
Eq. \ref{eq:average_code_length} with $l(i)$ as in Eq. \ref{eq:soft_optimal_length_uniquely_decodable_encoding}
yields the entropy of the rank distribution, i.e. 
\begin{equation}
\left< l \right>_{min}^{ud} = H = - \sum_i p(i) \log_N p(i).
\label{eq:entropy}
\end{equation}

The following property indicates that, when Zipf's law holds, the optimal code length under unique decodability is 
shifted linearly from
the optimal code length under non-singular coding. By the same token, $H$ is 
shifted linearly from
$\left<\log_N i \right>$. The same applies to the coding efficiencies.

\begin{property}
\label{label:separation_between_optima}
Zipf's rank-frequency law with exponent $\alpha$ and normalization factor $c$ is equivalent to a linear dependence between optimal code lengths with parameters $\alpha$ and $\beta = - \log_N c$ and $\alpha, \beta \geq 0$. First, $l_{min}^{ud}(i)$ and $\l_{min}^{ns}(i)$ are linearly dependent as
\begin{equation*}
l_{min}^{ud}(i) = \alpha \l_{min}^{ns}(i) + \beta.
\end{equation*}
Second, $\left< l \right>_{min}^{ud}= H$ and $\left< l \right>_{min}^{ns}=\left< \log_N i \right>$ are also linearly dependent, i.e. 
\begin{equation}
H = \alpha \left< \log_N i \right> + \beta.
\label{eq:linear_separation_of_expectations} 
\end{equation}
In addition, the coding efficiencies are linearly dependent as 
\begin{equation}
\eta^{ud} = \alpha \eta^{ns} + \frac{\beta}{\left< l \right>}.
\label{eq:linear_separation_of_efficiencies} 
\end{equation}
\end{property}
\begin{proof}
By taking logarithms (base $N$) on both sides of Eq. \ref{eq:Zipfs_rank_frequency_law}, one obtains
   \begin{eqnarray}
       \log_N p(i) & = & - \alpha \log_N i + \log_N c \label{eq:logarithmic_Zipfs_rank_frequency_law} \\
       \underbrace{-\log_N p(i)}_{l_{min}^{ud}(i)} & = & \alpha \underbrace{\log_N i}_{l_{min}^{ns}(i)} + \beta \nonumber \\
                                                   &   & \mbox{~(Eqs. \ref{eq:soft_optimal_length_non_singular_coding} and \ref{eq:soft_optimal_length_uniquely_decodable_encoding})} \label{eq:intermedidate}                   
   \end{eqnarray}   
where $\beta = - \log_N c$. Multiplying by $p(i)$ on both sides of Eq. \ref{eq:intermedidate} and summing over $i$, one obtains 
   \begin{eqnarray*}
       - \sum_i p(i) \log_N p(i) & = &  \alpha \sum_i p(i) \log_N i + \beta \sum_i p(i) \\
       H & = &  \alpha \left<\log_N i \right> + \beta \\
         &   & \mbox{ (\ref{eq:average_log_rank}, Eqs. \ref{eq:entropy} and $\sum_i p(i)= 1$)}       
   \end{eqnarray*}
Dividing both sides of the the last result by $\left< l \right>$, one obtains Eq. \ref{eq:linear_separation_of_efficiencies}. We have $\alpha \geq 0$ by definition of rank. To conclude that $\beta \geq 0$, one does not need to know the precise definition of $c$. Notice that $c = p(1)$ (Eq. \ref{eq:Zipfs_rank_frequency_law}). As $0 < p(1) \leq 1$, $\beta > 0$ follows.
\end{proof}

These properties illuminate the connection between Zipf's law and optimal coding but do not imply that systems obeying Zipf's law are coding optimally. 
To go further, we define a class of quasioptimal coding systems.

\begin{definition}
\label{def:new_class}
A class of coding systems such that
\begin{enumerate}
\item 
Coding can be optimal, that is \cite{Ferrer2019c} 
$$\tau(p(i), l(i)) \leq 0$$
\item
For each unit $i$, $l(i)$ is linearly displaced from optimal coding with respect to each of the two coding schemes, namely the code for unit $i$ satisfies the 
size-rank law \cite{Torre2019a}, i.e. 
\begin{eqnarray}
l(i) & = & a_{ns} l_{min}^{ns}(i) + b_{ns} \nonumber \\ 
     & = & a_{ns} \log_N i + b_{ns} \label{eq:linear_separation_from_optimal_non_singular_coding}
\end{eqnarray}
and the size-probability law, i.e. 
\begin{eqnarray} 
l(i) & = & a_{ud} l_{min}^{ud}(i) + b_{ud} \nonumber \\
     & = & -a_{ud} \log_N p(i) + b_{ud}, \label{eq:linear_separation_from_optimal_uniquely_decodable_encoding}
\end{eqnarray}
where $a_{ns}$, $a_{ud}$, $b_{ns}$ and $b_{ud}$ are constants. 
\end{enumerate}
\end{definition}
In that class, $a_{ns}$ and $b_{ns}$ are the parameters (slope and intercept) of the linear displacement from optimal coding non-singular ($ns$) coding whereas
$a_{ud}$ and $b_{ud}$ are the parameters of the linear displacement from optimal uniquely decodable ($ud$) coding. 
Notice that the size-rank law and the size-probability law are two possible translations of the law of abbreviation as a rank correlation (Eq. \ref{eq:law_of_abbreviation}) into a precise functional dependency between variables.
The next property shows that coding systems within that class obey Zipf's law.
\begin{property}
\label{prop:Zipfs_law_trap}
The coding systems in the class (Definition \ref{def:new_class}) satisfy 
\begin{equation*}
a_{ns}, a_{ud} \geq 0
\end{equation*}
and obey Zipf's law (Eq. \ref{eq:Zipfs_rank_frequency_law}) with
\begin{eqnarray}
\alpha & = & \frac{a_{ns}}{a_{ud}} \geq 0 \label{eq:slope_in_the_class}\\
c & = & N^{\frac{b_{ud} - b_{ns}}{a_{ud}}}. \nonumber
\end{eqnarray}  
\end{property}

\begin{proof}
The condition $\tau(p(i), l(i)) \leq 0$ of the class implies that $a_{ns}, a_{ud} \geq 0$. To see that $a_{ns} \geq 0$, notice that
$\tau(i, \log_N i) > 0$ by definition. Then $\tau(p(i), \log_N i) \leq 0$ as $p(i)$ is a non-increasing function of $i$. Finally 
\begin{equation*}
\tau(p(i), a_{ns} \log_N i + b_{ns}) = \tau(p(i), l(i)) \leq 0
\end{equation*} 
if and only if $a_{ns} \geq 0$. To see that $a_{ud} \geq 0$, notice that
$\tau(p(i), - \log_N p(i)) \leq 0$ by definition. Then 
\begin{equation*}
\tau(p(i), -a_{ud} \log_N p(i) + b_{ud}) = \tau(p(i), l(i)) \leq 0
\end{equation*}
if and only if $a_{ud} \geq 0$.

By equating the right hand side of Eqs. \ref{eq:linear_separation_from_optimal_non_singular_coding} and \ref{eq:linear_separation_from_optimal_uniquely_decodable_encoding}, one obtains
\begin{eqnarray*}
-a_{ud} \log_N p(i) + b_{ud} & = & a_{ns} \log_N i + b_{ns} \mbox{~(Eqs. \ref{eq:soft_optimal_length_non_singular_coding} and \ref{eq:soft_optimal_length_uniquely_decodable_encoding})} \\
\log_N p(i) & = & -\frac{a_{ns}}{a_{ud}} \log_N i + \underbrace{\frac{b_{ud} - b_{ns}}{a_{ud}}}_{\log_N c} \\
p(i) & = & ci^{-\alpha} \\ 
     &   & \mbox{ (exponentiating with base $N$)}
\end{eqnarray*}
with $\alpha$ and $c$ as in Eq. \ref{eq:slope_in_the_class}. $\alpha\geq 0$ follows from $a_{ns}$, $a_{ud} \geq 0$. 
\end{proof}

\section{The members of the class}

A critical question is if the class is empty. Indeed, this class contains at least random typing, namely pressing keys at random from a keyboard formed by the space and $N$ characters \cite{Mandelbrot1953,Conrad2004a}. In particular, the class contains a simple random typing model where the space has probability $p_s$, and the $N$ characters are equally likely \cite{Miller1957, Miller1963}. Recall that $N > 1$ and assume that empty strings cannot be generated (the space cannot be pressed more than once in a row). First, the probability that random typing produces a word of rank $i$ is \cite{Ferrer2019c}
\begin{equation}
p(i) = \frac{p_s}{1-p_s}\left(\frac{1-p_s}{N}\right)^{l(i)},
\label{eq:Zipfs_rank_frequency_law_random_typing}
\end{equation}
where $l(i)$ is the length of a word of probability rank $i$ according to optimal non-singular coding (Eq. \ref{eq:hard_optimal_length_non_singular_coding}).
As Eq. \ref{eq:hard_optimal_length_non_singular_coding} can be approximated by 
Eq. \ref{eq:soft_optimal_length_non_singular_coding}, random typing satisfies Eq. \ref{eq:linear_separation_from_optimal_non_singular_coding} with parameters $a_{ns} = 1$ and $b_{ns} = 0$. 
Second, random typing exhibits another specific form of the law of abbreviation, that follows from taking logs (base $N$) on Eq. \ref{eq:Zipfs_rank_frequency_law_random_typing}, that is \cite{Ferrer2019c}
\begin{equation}
l(i) = a_{RT} \log_N p(i) + b_{RT},
\end{equation}
where 
\begin{eqnarray*}
a_{RT} & = & \frac{1}{\log_N (1 - p_s) - 1} \\
b_{RT} & = & a_{RT} \log_N \frac{1 - p_s}{p_s}.
\end{eqnarray*}
Therefore, random typing satisfies Eq. \ref{eq:linear_separation_from_optimal_uniquely_decodable_encoding} with parameters $a_{ud} = - a_{RT}$ and $b_{ud} = b_{RT}$.
As a sanity check, let us verify that random typing satisfies Zipf's law theoretically as expected for a member of this class. Notice that Eq. \ref{eq:Zipfs_rank_frequency_law_random_typing} can be rewritten equivalently as
\begin{eqnarray*}
p(i) & \approx & \frac{p_s}{1-p_s}\left(N^{\log_N \frac{1 - p_s}{N}} \right)^{\log_N i} \mbox{ (Eq. \ref{eq:soft_optimal_length_non_singular_coding})}\\
     & = & \frac{p_s}{1-p_s}\left(N^{\log_N i} \right)^{\log_N \frac{1 - p_s}{N}} \\
     & = & c i^{-\alpha}
\end{eqnarray*}
with 
\begin{eqnarray*}
c      & = & \frac{p_s}{1-p_s} \\
\alpha & = & -\log_N \frac{1-p_s}{N}.
\end{eqnarray*}

Then the question is if human languages are also members of that class. It has been shown that Catalan, Spanish and English exhibit the law of abbreviation (Eq. \ref{eq:law_of_abbreviation}) and obey Eq. \ref{eq:linear_separation_from_optimal_non_singular_coding} and Eq. \ref{eq:linear_separation_from_optimal_uniquely_decodable_encoding} \cite{Hernandez2019b,Torre2019a,Corral2020a} 
\footnote{For Eq. \ref{eq:linear_separation_from_optimal_non_singular_coding} in English, there is a breakpoint separating two regimes requiring different values of $a_{ns}$ and $b_{ns}$. \cite{Corral2020a} found confirmation of Eq. 
\ref{eq:linear_separation_from_optimal_uniquely_decodable_encoding} for word lengths between 1 and 5. 
}. In addition, human languages that agree with Zipf's law (Eq. \ref{eq:Zipfs_rank_frequency_law}) \cite{Zipf1949a, Mehri2017a} are candidates for additional members of that class. 

As Zipf's law is a requirement to be a member of that class (Property \ref{prop:Zipfs_law_trap}), communication systems that exhibit an exponential distribution cannot be members of that class. Indeed, an exponential rank distribution has been found in ``key signs'' in rhesus monkeys \cite{Schleidt1973a}, ``tonemes'' in dolphins \cite{Dreher1961a} and calls in warbling vireos \cite{Howes-Jones1988a}. In the past, various authors concluded that ``tonemes'' in dolphins and calls in warbling vireos exhibit Zipf's law based on a straight line in logarithmic scale for frequency and normal scale for frequency rank \cite{Janik2006a,Dreher1961a,Howes-Jones1988a}. However, this is indeed evidence for an exponential distribution. Such a confusion has been reviewed and discussed recently \cite{Ferrer2024b}. Finally, notice that evidence in dolphins is mixed because Zipf's law has been found in dolphin vocalizations \cite{Markov1990a} and also specifically in dolphin whistles \cite{McCowan1999,McCowan2005a}. Therefore, as dolphin whistles agree with Zipf's law of abbreviation \cite{Vradi2021a}, dolphins could be members of that class. The case of humpback whale song is similar. Although Zipf's law of abbreviation has been found in humpback whale song \cite{Arnon2025a}, evidence of Zipf's law is heterogeneous \cite{Allen2019a, Arnon2025a}.

\section{A group-like structure}

The three laws that are satisfied by members of the new class, i.e. the size-rank law (Eq. \ref{eq:linear_separation_from_optimal_non_singular_coding}), the size-probability law (Eq. \ref{eq:linear_separation_from_optimal_uniquely_decodable_encoding}) and Zipf's law, that is a probability-rank law (Eq. \ref{eq:Zipfs_rank_frequency_law}), form a group-like structure. 
In particular, these three equations form a set where every pair of them yields the equation that is left. We have shown how the size-rank law and the size-probability law yield Zipf's law (Property \ref{prop:Zipfs_law_trap}). As already argued by \cite{Torre2019a}, Zipf's law and what we call the size-probability law give the size-rank law. In particular, the application of Eq. \ref{eq:logarithmic_Zipfs_rank_frequency_law} to Eq. \ref{eq:linear_separation_from_optimal_uniquely_decodable_encoding} yields Eq. \ref{eq:linear_separation_from_optimal_non_singular_coding} with 
\begin{eqnarray*}
a_{ns} & = & \alpha a_{ud} \\
b_{ns} & = & b_{ud} - a_{ud} \log_N c.
\end{eqnarray*}
Finally, Zipf's law and the size-rank law yield the size-probability law. In particular, Eq. \ref{eq:logarithmic_Zipfs_rank_frequency_law} is equivalent to 
\begin{equation*}
\log_N i = \frac{\log_N c - \log_N p(i)}{\alpha},
\end{equation*}
which transforms Eq. \ref{eq:linear_separation_from_optimal_non_singular_coding} into Eq. \ref{eq:linear_separation_from_optimal_uniquely_decodable_encoding} with 
\begin{eqnarray*}
a_{ud} & = & \frac{a_{ns}}{\alpha} \\
b_{ud} & = & b_{ns} + \frac{a_{ns}\log_N c}{\alpha}.
\end{eqnarray*}

\section{Discussion}

\subsection{A new interpretation of rank distributions in log-log scale}

Our information theoretic framework provides a new insight into plots of frequency versus rank in double logarithmic scale: the appearance of a straight line indicates (linear) proximity between the lengths of two kinds of optimal codes for a word type given its frequency and its rank (Property \ref{label:separation_between_optima}). The two kinds of optimal codes are the non-singular codes and a subclass that are the uniquely decodable codes. The exponent of Zipf's law is the slope of the linear transformation that links the optimal lengths. Such an insight does not make any assumption: it just follows from the mathematical shape of Zipf's law and the optimal lengths of the codes.
In a linguistically diverse sample of languages, $\alpha$ ranges from 0.765 in Thai to 1.442 in Korean and has an average value of 1.130 \cite{Mehri2017a}. 
When controlling for linguistic family, $\bar{\alpha}$, the average $\alpha$ within a linguistic family, exhibits a similar range of variation, i.e. $\bar{\alpha} \in [0.765, 1.357]$. 
 
A more powerful insight can be obtained by assuming that the system is compressing, namely reducing the length of codes. In that case, the straight line in log-log scale would indicate proximity of the system to optimal coding. If the gap between the length of words and the corresponding optimal length is only linear, then the system must exhibit Zipf's law (Property \ref{prop:Zipfs_law_trap}). Languages (at least Catalan, Spanish and English) exhibit such a linear proximity, which is consistent with a strong form of compression. If one assumes that all languages belong to that class, one finds that $\bar{\alpha} > 1$ in 23 out of 28 families \cite{Mehri2017a}, suggesting that $a_{ns} > a_{ud}$ in most families and that $a_{ns} \approx 1.13 a_{ud}$ on average across languages (recall Eq. \ref{eq:slope_in_the_class}). The actual values of $a_{ns}$ and $a_{ud}$ in a diverse sample of languages should be the subject of future research.
The failure to find Zipf's law in certain species but an exponential distribution instead \cite{Schleidt1973a,Howes-Jones1988a,Dreher1961a} suggests that pressure for compression may not be strong enough for these species or that we are not looking at the right units. In humans, part-of-speech tags, word orders, colors, kinship terms, and verbal alternation classes also exhibit an exponential distribution \cite{Tuzzi2010a, Cysouw2010a, Ferrer2024b, Ramscar2019a}. 

\subsection{A new class of coding optimality}

We presented the hypothesis that languages belong to a class of communication systems that are linearly displaced from these two kinds of optimal coding and thus must exhibit Zipf's law. This is consistent with the view of Zipf's law as a characteristic of the most compressed lossless representations \cite{Marsili2022a}. We have confirmed that the class contains at least three languages. The ubiquity of Zipf's law in languages \cite{Zipf1949a,Mehri2017a} suggests that this class would comprise many more languages. We cannot rule out the possibility that it does not cover all. A statistical analysis of a wider set of languages should be the target of future research. 

At present, the only other species that could be a member of this class are dolphins and humpback whales. Although the evidence of Zipf's law in dolphins is mixed with the exponential distribution \cite{Dreher1961a,McCowan1999,McCowan2005a}, humans also exhibit such a kind of mixed evidence \cite{Zipf1949a,Tuzzi2010a,Ramscar2019a}. Differences in sampling, segmentation or categorization criteria may explain why Zipf's law has not always been found in dolphin whistles. Indeed, variations in segmentation methods impact on the quality of the agreement with Zipf's law in humpback whales \cite{Allen2019a,Arnon2025a}. The presence of non-adjacent dependencies in dolphin whistle sequences rules out random typing as an explanation for the presence of Zipf's law in dolphins \cite{Ferrer2012c}.

In spite of the considerations above, the class imposes no strong conditions on the parameters of the linear displacement from optimality ($a_{ns}$, $b_{ns}$, $a_{ud}$ and $b_{ud}$). Thus the displacement from optimality could be large but still linear. If that happened, the question would be: how could a system be truly far from optimality but displaying functional dependencies (the size-rank law and size-probability law) whose mathematical structure follows from optimal coding assumptions? In fact, languages are not very far from optimality. The analysis of $\eta$ in a sample of more than 1000 languages from the Parallel Bible Corpus (BPC) \cite{Mayer2014}, revealed that, on average across languages, $\eta^{ns} = 30\%$ and $\eta^{ud} = 40\%$ \cite{Ferrer2018b}. The proximity between $\eta^{ns}$ and $\eta^{ud}$ is likely to stem from Zipf's law (Eq. \ref{eq:linear_separation_of_efficiencies}) and the  narrow range of variation of $\alpha$ \cite{Mehri2017a}. For a member of that class, the exponent $\alpha$ is the ratio between the slopes of the linear displacement between the actual code length and the optimal length (Eq. \ref{eq:slope_in_the_class}) and such a ratio is bounded in human languages \cite{Mehri2017a}.   

Finally notice that displacement of languages from optimal coding may have two reasons. First, it may reflect the simplifications of our coding framework, e.g., that the $N$ symbols of the alphabet used to compose codes have the same cost, a hidden assumption of standard information theory \cite{Borda2011a}. Second, it may reflect the inability of languages to reach the theoretical optimum. The nature of such a gap should be the subject of further research.

\subsection{The origins of Zipf's law}

An elementary question is if Zipf's law is inevitable due to the definition of the variables involved. Recently, it has been shown that any rank distribution satisfies \cite{Debowski2025a}  
\begin{equation*}
p(i) \leq \frac{1}{i},
\end{equation*}
which implies that rank distributions are channeled to show a power-law (Eq. \ref{eq:Zipfs_rank_frequency_law}) with $\alpha = 1$. However, we have reviewed above multiple real examples of systems that exhibit an exponential rank distribution. Another question is if random typing could be the reason why human languages and other systems show Zipf's law. In the past, random typing has been regarded as a proof that Zipf's law can be retrieved without involving any optimization \cite{Miller1957, Miller1963, Li1998, Suzuki2004a}. That argument is flawed because random typing is an optimal non-singular coding system. However, one can still retain that random typing is a proof that Zipf's law is easy to retrieve by virtue of some simple stochastic process. However, this raises the question of why certain species do not exhibit Zipf's law for word frequencies. The only answer that random typing provides is that $N = 1$ yields an exponential distribution while $N > 1$ yields a rank distribution and a frequency spectrum whose ability to mimic the actual distribution of word frequencies has been criticized heavily \cite{Ferrer2009b,Ferrer2009a}. Random typing with equal letter probabilities fails to cover the actual range of variation of $\alpha$, e.g., \cite{Mehri2017a}, because it only reproduces $\alpha \geq 1$ \cite{Ferrer2004e}. \footnote{It is well-known that $\alpha = - \log_N (1 - p_s) + 1 \geq 1$ \cite{Conrad2004a}. Notice that such a result can also be retrieved from $\alpha = a_{ns}/a_{ud} = -1/a_{RT}$. A discussion of the potential range of variation of the exponent covered by random typing with unequal character probabilities \cite{Conrad2004a,Bochkarev2017a} is beyond the scope of this letter.        
} 
  
In contrast, optimal coding approaches shed light on the origins of both the law of abbreviation, Zipf's law and the alternative exponential \cite{Mandelbrot1966,Ferrer2016b,Ferrer2019c}. 
The simplest and most ancient optimal coding argument for the origins of Zipf's law is Mandelbrot maximum entropy (maxent) model \cite{Mandelbrot1966}. The core of Mandelbrot's argument is the optimization of the functional
\begin{equation*}
J = H + \alpha \left<l\right> + \beta \sum_{i} p_i
\end{equation*}
where $\alpha$ and $\beta$ are Lagrange multipliers. 
If one assumes optimal non-singular coding (with $N \gg 1)$, then $\left<l\right>$ becomes $\left< \log_N i\right>$ and the solution to the {\em maxent} problem is Zipf's law (Eq. \ref{eq:Zipfs_rank_frequency_law}) as already pointed out by Mandelbrot \cite{Mandelbrot1966}.
If one assumes $N = 1$ or replaces $\left<l\right>$ by the expected rank, that is simply, $\left<i\right>$, it is well-known that the solution will be an exponential distribution \cite{Kapur1992a}.
Mandelbrot's classic argument raises two concerns. The first one is the validity of the {\em maxent} argument for non-equilibrium systems such as language. One can argue that the {\em maxent} still provides the most unbiased distribution satisfying given constraints \cite{Kapur1992a}.
The second one is the validity of the assumption of optimal non-singular coding. Recall that $\eta^{ns}$ is on average $30\%$ \cite{Ferrer2018b}. Here we have put forward a class of languages that overcomes the limitations of Mandelbrot's argument. The members of that class do not need to be coding optimally for Zipf's law to surface, they only need to be linearly close to the optima. 
Modern optimization models about the origins of Zipf's law have similar limitations \cite{Ferrer2004e,Marsili2022a}. For instance, the proposal of a general relevance-resolution trade-off \cite{Marsili2022a} (see Appendix A of \cite{Bellina2025a} for a language-oriented summary of the framework) suggests, at first glance, an alternative general explanation to the origins of Zipf's law that does not involve word lengths as ours. 
However, it assumes optimal uniquely decodable encoding when defining speaker and hearer costs and in particular when defining the cost of a word. Recall that $\eta^{ns}$ is on average $40\%$ \cite{Ferrer2018b}. In contrast, our proposal does not impose either optimal coding or a specific coding scheme. Instead, it uses optimal non-singular coding and optimal uniquely decodable encoding as reference points from which natural languages are hypothesized to deviate linearly from. While we aim to provide a unified explanation to linguistic laws via compression in communication systems, \cite{Marsili2022a} has a much wider scope in terms of systems. We hope that our research stimulates a dialogue between these two complementary approaches. 

Optimal coding models shed light into why some systems do not get to Zipf's law. Only coding systems that compress strongly under suitable constraints can get close to the region of the space of possible systems where they are trapped by Zipf's law (Property \ref{prop:Zipfs_law_trap}). If our hypothesis is correct, coding systems where unique segmentation (unique decodability) is not relevant would be less likely to exhibit Zipf's law. A simple example are systems that produce calls that are separated by silences and then achieving unique segmentation is trivial or systems that produce isolated calls or continuous sequences of calls that are short enough for pressure for unique segmentation to be irrelevant. As for the latter, the vocal sequences of a wide range of primates are short \cite{Girard-Buttoz2022a} and thus investing in unique segmentation may not be worth it for them or the lack of design for unique segmentation may prevent them from getting to longer sequences.
To sum up, other species are telling us that reproducing Zipf's law for word frequencies is not as easy as commonly believed \cite{Miller1963, Suzuki2004a} and optimal coding yields a unified explanation to Zipf's law of abbreviation, Zipf's rank frequency law and the conditions required for the emergence of Zipf's law in communication systems. 

Inferring just from Zipf's law the right model is an impossible mission \cite{Suzuki2004a}. Thus other aspects of the systems under consideration must be examined and model selection must be guided by parsimony and explanatory power. The hypothesis of compression \cite{Mandelbrot1966,Ferrer2015b,Ferrer2016b,Ferrer2019c,Marsili2022a}
offers a parsimonious explanation to the four linguistic laws above: the law of abbreviation, its two specific forms (the size-rank law and the size-probability law) and Zipf's law. Furthermore, if one assumes that compression is the reason for the origins of Zipf's law of abbreviation, the robust support for that pattern across languages \cite{Sanada2008a,Bentz2016a,Wang2015a,Meylan2021a,Petrini2022b,Koshevoy2023a} 
turns compression as the strongest candidate for the origins of Zipf's law in languages at present.  
The key point is whether Zipf's law is a cause or a consequence. We have shown that Zipf's law is one of three laws that form a group-like structure. Given two of the laws, the other one follows. Thus, within the new class of optimality, Zipf's law could be both a consequence or a cause. Here we have argued for Zipf's law as a consequence of compression in languages. 

\acknowledgments
This research was presented during the annual meeting of the Catalan Association for the Study of Complex Systems (May 16, 2025).
We are grateful to R. Pastor-Satorras for helpful comments and discussions.  
This research is supported by a recognition 2021SGR-Cat (01266 LQMC) from AGAUR (Generalitat de Catalunya) and the grant AGRUPS-2025 from Universitat Politècnica de Catalunya.

\bibliographystyle{eplbib}
\bibliography{../../../../Dropbox/biblio/rferrericancho.bib,../../../../Dropbox/biblio/complex.bib, ../../../../Dropbox/biblio/ling.bib, ../../../../Dropbox/biblio/cl.bib, ../../../../Dropbox/biblio/cs.bib, ../../../../Dropbox/biblio/maths.bib}

\begin{thebibliography}{10}
\expandafter\ifx\csname url\endcsname\relax\def\url#1{\texttt{#1}}\fi

\bibitem{Leben2018a}
\Name{Leben W.~R.} \Book{Languages of the world} (Oxford University Press)
  2018.

\bibitem{Zipf1949a}
\Name{Zipf G.~K.} \Book{Human behaviour and the principle of least effort}
  (Addison-Wesley, Cambridge (MA), USA) 1949.

\bibitem{Bentz2016a}
\Name{Bentz C. \and {Ferrer-i-Cancho} R.} \Book{{Zipf's law of abbreviation as
  a language universal}} presented at \Book{{Proceedings of the Leiden Workshop
  on Capturing Phylogenetic Algorithms for Linguistics}}, edited by \Name{Bentz
  C., J{\"a}ger G. \and Yanovich I.} (University of T{\"u}bingen, Leiden, The
  Netherlands) 2016.

\bibitem{Sanada2008a}
\Name{Sanada H.} \Book{Investigations in Japanese historical lexicology} (Peust
  \& Gutschmidt Verlag, G\"{o}ttingen) 2008.

\bibitem{Wang2015a}
\Name{Wang Y. \and Chen X.} \Book{Structural complexity of simplified {Chinese}
  characters} in \Book{Recent Contributions to Quantitative Linguistics},
  edited by \Name{Tuzzi A. \and M.~Benesov\'a J.~M.} (De Gruyter) 2015 pp.
  229--239.

\bibitem{Meylan2021a}
\Name{Meylan S.~C. \and Griffiths T.~L.} \REVIEW{Cognitive
  Science}{45}{2021}{e12983}.

\bibitem{Petrini2022b}
\Name{Petrini S., {Casas-i-Muñoz} A., {Cluet-i-Martinell} J., Wang M., Bentz
  C. \and {Ferrer-i-Cancho} R.} \REVIEW{Glottometrics}{54}{2023}{58}.

\bibitem{Koshevoy2023a}
\Name{Koshevoy A., Miton H. \and Morin O.}
  \REVIEW{Cognition}{238}{2023}{105527}.

\bibitem{Semple2021a}
\Name{Semple S., {Ferrer-i-Cancho} R. \and Gustison M.} \REVIEW{Trends in
  Ecology and Evolution}{37}{2022}{53}.

\bibitem{Ferrer2019c}
\Name{{Ferrer-i-Cancho} R., Bentz C. \and Seguin C.} \REVIEW{Journal of
  Quantitative Linguistics}{29}{2022}{165}.

\bibitem{Cover2006a}
\Name{Cover T.~M. \and Thomas J.~A.} \Book{Elements of information theory}
  (Wiley, New York) 2006 2nd edition.

\bibitem{Elias1975a}
\Name{{Elias} P.} \REVIEW{IEEE Transactions on Information
  Theory}{21}{1975}{194}.

\bibitem{Mandelbrot1966}
\Name{Mandelbrot B.} \Book{Information theory and psycholinguistics: a theory
  of word frequencies} in \Book{Readings in mathematical social sciences},
  edited by \Name{Lazarsfield P.~F. \and Henry N.~W.} (MIT Press, Cambridge)
  1966 pp. 151--168.

\bibitem{Shannon1948}
\Name{Shannon C.~E.} \REVIEW{Bell Systems Technical Journal}{27}{1948}{379}.

\bibitem{Hernandez2019b}
\Name{Hernández-Fernández A., G.~Torre I., Garrido J.-M. \and Lacasa L.}
  \REVIEW{Entropy}{21}{2019}{}.

\bibitem{Torre2019a}
\Name{Torre I.~G., Luque B., Lacasa L., Kello C.~T. \and Hernández-Fernández
  A.} \REVIEW{Royal Society Open Science}{6}{2019}{191023}.

\bibitem{Mehri2017a}
\Name{Mehri A. \and Jamaati M.} \REVIEW{Physics Letters A}{381}{2017}{2470}.

\bibitem{Borda2011a}
\Name{Borda M.} \Book{Fundamentals in information theory and coding} (Springer,
  Berlin) 2011.

\bibitem{Mandelbrot1953}
\Name{Mandelbrot B.} \Book{An informational theory of the statistical structure
  of language} in \Book{Communication theory}, edited by \Name{Jackson W.}
  (Butterworths, London) 1953 pp. 486--502.

\bibitem{Conrad2004a}
\Name{Conrad B. \and Mitzenmacher M.} \REVIEW{IEEE Transactions on Information
  Theory}{50}{2004}{1403}.

\bibitem{Miller1957}
\Name{Miller G.~A.} \REVIEW{Am. J. Psychol.}{70}{1957}{311}.

\bibitem{Miller1963}
\Name{Miller G.~A. \and Chomsky N.} \Book{Finitary models of language users} in
  \Book{Handbook of Mathematical Psychology}, edited by \Name{Luce R.~D., Bush
  R. \and Galanter E.} Vol.~2 (Wiley, New York) 1963 pp. 419--491.

\bibitem{Corral2020a}
\Name{Corral A. \and Serra I.} \REVIEW{Entropy}{22}{2020}{}.

\bibitem{Schleidt1973a}
\Name{Schleidt W.~M.} \REVIEW{Journal of Theoretical Biology}{42}{1973}{359}.

\bibitem{Dreher1961a}
\Name{Dreher J.~J.} \REVIEW{The Journal of the Acoustical Society of
  America}{33}{1961}{1799}.

\bibitem{Howes-Jones1988a}
\Name{Howes-Jones D. \and Barlow J.} \Book{The structure of the call note
  system of the warbling vireo} in \Book{Royal Ontario Museum, Life Sciences
  Contributions} Vol. 151 (Royal Ontario Museum, Toronto) 1988.

\bibitem{Janik2006a}
\Name{Janik V.} \Book{Communication in marine mammals} in \Book{Encyclopedia of
  Language \& Linguistics}, edited by \Name{Brown K.} second edition Edition
  (Elsevier, Oxford) 2006 pp. 646--654.

\bibitem{Ferrer2024b}
\Name{{Ferrer-i-Cancho} R.}
  \REVIEW{https://arxiv.org/abs/2502.06342}{}{2024}{}.

\bibitem{Markov1990a}
\Name{Markov V.~I. \and Ostrovskaya V.~M.} \Book{Organization of communication
  system in {Tursiops truncatus Montagu}} in \Book{Sensory abilities of
  cetaceans - Laboratory and Field Evidence}, edited by \Name{Thomas J.~A. \and
  Kastelein R.~A.} Vol. 196 of \emph{NATO ASI Series. Series A: Life Sciences}
  (Plenum Press, New York) 1990 pp. 599--602.

\bibitem{McCowan1999}
\Name{McCowan B., Hanser S.~F. \and Doyle L.~R.} \REVIEW{Animal
  Behaviour}{57}{1999}{409}.

\bibitem{McCowan2005a}
\Name{McCowan B., Doyle L.~R., Jenkins J.~M. \and Hanser S.~F.} \REVIEW{Animal
  Behaviour}{69}{2005}{F1}.

\bibitem{Vradi2021a}
\Name{Vradi A.} \Book{Dolphin communication: a quantitative linguistics
  approach} Master's thesis Barcelona School of Informatics Barcelona (2021).

\bibitem{Arnon2025a}
\Name{Arnon I., Kirby S., Allen J.~A., Garrigue C., Carroll E.~L. \and Garland
  E.~C.} \REVIEW{Science}{387}{2025}{649}.

\bibitem{Allen2019a}
\Name{Allen J.~A., Garland E.~C., Dunlop R.~A. \and Noad M.~J.}
  \REVIEW{Proceedings of the Royal Society B: Biological
  Sciences}{286}{2019}{20192014}.
\newline\url{http://dx.doi.org/10.1098/rspb.2019.2014}

\bibitem{Tuzzi2010a}
\Name{Tuzzi A., Popescu I.-I. \and Altmann G.} \Book{Quantitative analysis of
  {Italian} texts} Vol.~6 of \emph{Studies in Quantitative Linguistics} (RAM
  Verlag, L{\"u}denscheid, Germany) 2010.

\bibitem{Cysouw2010a}
\Name{Cysouw M.} \REVIEW{Linguistic Typology}{14}{2010}{253}.

\bibitem{Ramscar2019a}
\Name{Ramscar M.} \REVIEW{https://psyarxiv.com/e3hps}{}{2019}{}.

\bibitem{Marsili2022a}
\Name{Marsili M. \and Roudi Y.} \REVIEW{Physics Reports}{963}{2022}{1}.

\bibitem{Ferrer2012c}
\Name{{Ferrer-i-Cancho} R. \and McCowan B.} \REVIEW{Journal of Statistical
  Mechanics}{}{2012}{P06002}.

\bibitem{Mayer2014}
\Name{Mayer T. \and Cysouw M.} \Book{Creating a massively parallel {Bible}
  corpus} presented at \Book{Proceedings of the Ninth International Conference
  on Language Resources and Evaluation ({LREC}'14)}, edited by \Name{Calzolari
  N., Choukri K., Declerck T., Loftsson H., Maegaard B., Mariani J., Moreno A.,
  Odijk J. \and Piperidis S.} (Reykjavik, Iceland) 2014.

\bibitem{Ferrer2018b}
\Name{{Ferrer-i-Cancho} R. \and Bentz C.} \Book{The evolution of optimized
  language in the light of standard information theory} presented at \Book{The
  Evolution of Language: Proceedings of the 12th International Conference
  (EVOLANGXII)}, edited by \Name{Cuskley C., Flaherty M., Little H., McCrohon
  L., Ravignani A. \and Verhoef T.} (NCU Press) 2018.
\newline\url{http://evolang.org/torun/proceedings/papertemplate.html?p=142}

\bibitem{Debowski2025a}
\Name{Debowski L.} \REVIEW{Journal of Quantitative Linguistics}{32}{2025}{128}.

\bibitem{Li1998}
\Name{Li W.} \REVIEW{Complexity}{3}{1998}{9} letters to the editor.

\bibitem{Suzuki2004a}
\Name{Suzuki R., Tyack P.~L. \and Buck J.} \REVIEW{Animal
  Behaviour}{69}{2005}{9}.

\bibitem{Ferrer2009b}
\Name{{Ferrer-i-Cancho} R. \and Elvev{\aa}g B.} \REVIEW{PLoS
  ONE}{5}{2009}{e9411}.

\bibitem{Ferrer2009a}
\Name{{Ferrer-i-Cancho} R. \and Gavald\`a R.} \REVIEW{Journal of the American
  Association for Information Science and Technology}{60}{2009}{837}.

\bibitem{Ferrer2004e}
\Name{{Ferrer-i-Cancho} R.} \REVIEW{European Physical Journal
  B}{47}{2005}{449}.

\bibitem{Bochkarev2017a}
\Name{Bochkarev V., Lerner E., Nikiforov A. \and Pismenskiy A.} \REVIEW{Journal
  of Physics: Conference Series}{936}{2017}{012028}.

\bibitem{Ferrer2016b}
\Name{{Ferrer-i-Cancho} R.} \REVIEW{Complexity}{21}{2016}{409}.

\bibitem{Kapur1992a}
\Name{Kapur J.~N. \and Kesavan H.~K.} \Book{Entropy optimization principles and
  their applications} in \Book{Entropy and Energy Dissipation in Water
  Resources}, edited by \Name{Singh V.~P. \and Fiorentino M.} (Springer
  Netherlands, Dordrecht) 1992 pp. 3--20.

\bibitem{Bellina2025a}
\Name{Bellina A. \and Servedio V. D.~P.} \REVIEW{}{}{2025}{}.
\newline\url{https://arxiv.org/abs/2503.17512}

\bibitem{Girard-Buttoz2022a}
\Name{Girard-Buttoz C., Zaccarella E., Bortolato T., Friederici A.~D., Wittig
  R.~M. \and Crockford C.} \REVIEW{Communications Biology}{5}{2022}{}.
\newline\url{http://dx.doi.org/10.1038/s42003-022-03350-8}

\bibitem{Ferrer2015b}
\Name{{Ferrer-i-Cancho} R.} \REVIEW{Journal of Quantitative
  Linguistics}{25}{2018}{207}.

\end{thebibliography}
  
\end{document}